\documentclass[10pt,twocolumn,letterpaper]{article}

\usepackage{iccv}
\usepackage{times}
\usepackage{epsfig}
\usepackage{graphicx}
\usepackage{amsmath}
\usepackage{amssymb}
\usepackage{booktabs}
\usepackage{subcaption}
\usepackage[ruled,vlined]{algorithm2e}
\usepackage{multirow}
\usepackage{mathabx}
\usepackage{xcolor}
\usepackage{authblk}
\usepackage{algorithmic}

\usepackage[breaklinks=true,bookmarks=false]{hyperref}

\iccvfinalcopy % *** Uncomment this line for the final submission

\def\iccvPaperID{****} % *** Enter the ICCV Paper ID here

\ificcvfinal\fi

\begin{document}

%%%%%%%%% TITLE
\title{AutoAdapt: Automated Segmentation Network Search \\for Unsupervised Domain Adaptation}

% \author{Xueqing Deng\\
% UC Merced\\
% {\tt\small xdeng7@ucmerced.edu}
% % For a paper whose authors are all at the same institution,
% % omit the following lines up until the closing ``}''.
% % Additional authors and addresses can be added with ``\and'',
% % just like the second author.
% % To save space, use either the email address or home page, not both
% \and
% Yuxin Tian\\
% UC Merced\\
% {\tt\small ytian8@ucmerced.edu}

% \and
% Yi Zhu\\
% Amazon\\
% {\tt\small yzaws@amazon.com}

% \and
% Shawn Newsam\\
% UC Merced\\
% {\tt\small snewsam@ucmerced.edu}
% }

\author[1]{Xueqing Deng}
\author[2]{Yi Zhu} 
\author[1]{Yuxin Tian}
\author[1]{Shawn Newsam}
\affil[1]{EECS, UC Merced \textit {\{xdeng7, ytian8, snewsam\}@ucmerced.edu}}
\affil[2]{Amazon Web Services \textit {\{yzaws\}@amazon.com}}

\maketitle
% Remove page # from the first page of camera-ready.
\ificcvfinal\thispagestyle{empty}\fi

%%%%%%%%% ABSTRACT
\begin{abstract}
Neural network-based semantic segmentation has achieved remarkable results when large amounts of annotated data are available, that is, in the supervised case. However, such data is expensive to collect and so methods have been developed to adapt models trained on related, often synthetic data for which labels are readily available. Current adaptation approaches do not consider the dependence of the generalization/transferability of these models on network architecture. In this paper, we perform neural architecture search (NAS) to provide architecture-level perspective and analysis for domain adaptation. We identify the optimization gap that exists when searching architectures for unsupervised domain adaptation which makes this NAS problem uniquely difficult. We propose bridging this gap by using maximum mean discrepancy and regional weighted entropy to estimate the accuracy metric. Experimental results on several widely adopted benchmarks show that our proposed AutoAdapt framework indeed discovers architectures that improve the performance of a number of existing adaptation techniques. 

\end{abstract}

% %%%%%%%%% BODY TEXT

\section{Introduction}

Fully connected convolutional neural networks have shown to be effective for per-pixel classification (semantic segmentation) in images particularly in the supervised case~\cite{chen2017deeplabv2, chen2018deeplabv3plus, fu2019danet, zhao2017pspnet}. However, performance degrades when the networks are applied to domains which they were not trained on, that is, when there is a domain shift. Labeling data in the new domain is expensive so researchers have explored adapting the networks in an unsupervised manner~\cite{hoffman2016fcns, hoffman2018cycada}, a technique known as unsupervised domain adaptation (UDA).

\begin{figure}[htbp]
\includegraphics[width=\linewidth]{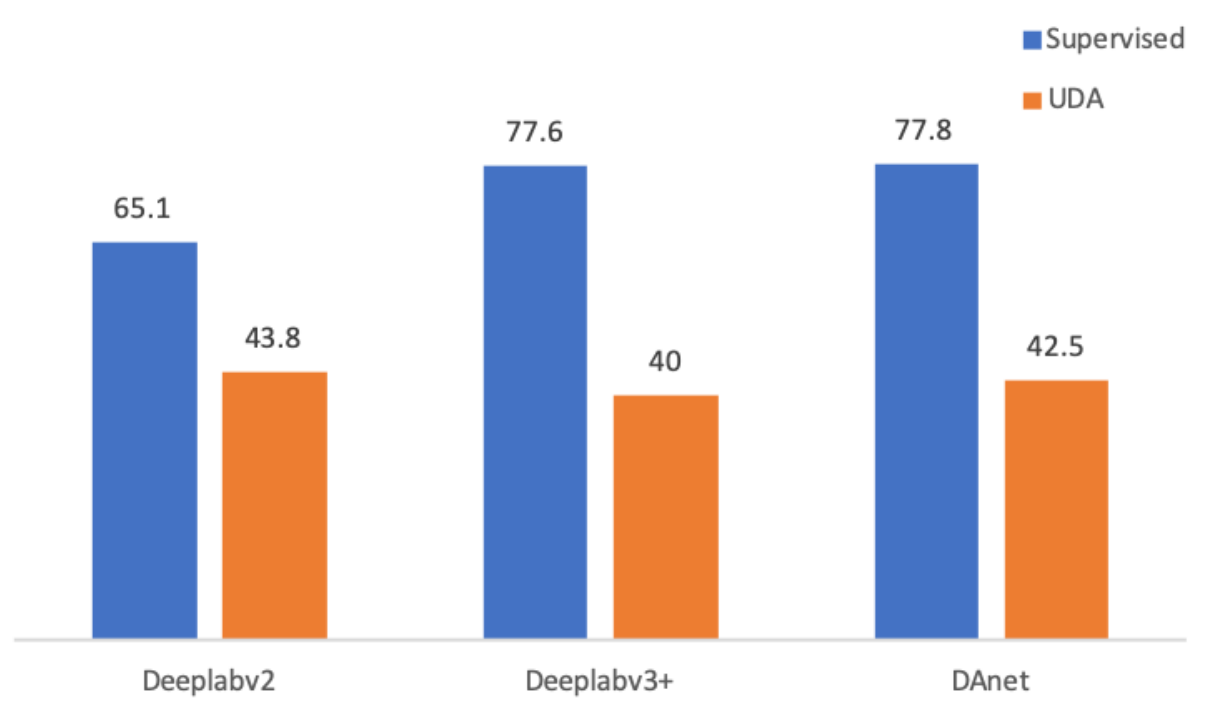}
\caption{Segmentation models that are better in the supervised case are not necessarily better for UDA. Separate architecture design must be done for UDA which motivates our automated approach AutoAdapt.}
%%%Comparison study of semantic segmentation model performance in supervised manner and UDA. Reversed improvement are made when the models are used in UDA.}
\label{fig:overview}
\end{figure}

UDA seeks to narrow the domain gap between a source dataset, for which we have images and labels, and a target dataset, for which we have only images. While good progress has been made in UDA for image classification~\cite{long2015dan, long2016unsupervised, tzeng2014ddc}, UDA for semantic segmentation remains a challenge since knowledge of both the image features and the image structure is needed in order to adapt.

%%%UDA generally seeks to narrow the domain gap between a source dataset, for which we have ample images with annotations, and a target dataset, for which we have only images. While good progress has been made in UDA for image classification~\cite{long2015dan, long2016unsupervised, tzeng2014ddc}, UDA for semantic segmentation remains a challenge since knowledge of both image features and image structure is needed in order to adapt.

Work has been proposed to address UDA at the feature-level~\cite{hoffman2016fcns}, at the image-level~\cite{hoffman2018cycada, zou2018cbst, chen2018road} and at the output space-level~\cite{tsai2018adaptseg, vu2019advent, pan2020intrada, li2020ccm}. However, we find some interesting observations at the architecture-level, in particular the lack of correlation between how well a model performs on a supervised problem and how well it performs for UDA. As shown in Fig.~\ref{fig:overview}, DeeplabV3+~\cite{chen2018deeplabv3plus} and DANet~\cite{fu2019danet} both outperform DeeplabV2~\cite{chen2017deeplabv2} on supervised semantic segmentation but perform worse in the UDA case. Model improvements in the supervised case do not necessarily translate to improvements for UDA. Model search must be performed separately for UDA but doing this manually is a significant undertaking especially since the optimal model can be dataset dependent. We therefore turn to the burgeoning field of neural architecture search (NAS). We design a search space over semantic segmentation models for UDA that includes advanced modules and configurations. We then develop a search framework that finds effective models that can be used with existing UDA approaches.

%%%Work has been proposed to address UDA at the feature-level~\cite{hoffman2016fcns}, at the image-level~\cite{hoffman2018cycada, zou2018cbst, zou2018crst, chen2018road} and at the output space-level~\cite{tsai2018adaptseg, vu2019advent, pan2020intrada, li2020ccm}. However, we find some interesting observations at the architecture-level. In our experiments shown in Fig.~\ref{fig:overview}, DeeplabV3+\cite{chen2018deeplabv3plus} is not performing as well as DeeplabV2\cite{chen2017deeplabv2} in UDA which contradicting the observations in supervised semantic segmentation. Similar observations are made for DANet\cite{fu2019danet}. This implies some architectures are not providing benefits for transfer learning/generalization/domain adaptation. In other words, simply borrowing architectures from supervised semantic segmentation doesn't work for UDA even decreases the adaptation performance. This brings the challenges to manually design UDA techniques, particularly when people try to use a new model for different datasets. Motivated by this,  we therefore raise the question "what architectures can provide generalization beneficial for adaptation?". Rather than manually testing whether the large space of innovative architectures can improve UDA, we turn to the burgeoning field of neural architecture search (NAS). We seek to design a search space over semantic segmentation models for UDA that includes advanced modules and configurations and then develop a framework to find an effective model that should work for most UDA techniques.

However, as we will point out, performing automated search over candidate model architectures for UDA including different training and adaptation configurations is a difficult problem. Simply combining existing NAS search methods with UDA approaches is not effective. Tighter integration between these two concepts is needed particularly regarding how to evaluate candidate architectures when labeled data is not available for the target domain. We study the optimization gap that results when using NAS for UDA and bridge this gap by simulating the target domain evaluation metric. We incorporate these contributions into a complete search framework.

%%%However, as we point out, performing automated search over candidate model architectures for UDA including different training and adaptation configurations is not a simple problem. We show that simply combining existing NAS search methods with UDA approaches is not effective. A tighter integration of these two concepts presents challenges such as how to evaluate candidate architectures when there is no labeled data in the target domain. This results in an optimization gap during searching. In this paper, we first analyze what is the optimization gap unique for UDA. Then we propose to overcome the gap by simulate the evaluation metric that with labels. Finally, we combine the search strategy to propose the overall framework.

% Also, architecture search can be very expensive especially when the search space is large and so efficiency needs to be considered.

We make the following contributions in this work:

\begin{itemize}

    \item We approach UDA from a novel perspective, namely \textit{architecture}. We propose AutoAdapt, the first framework to our knowledge to perform automated network search for UDA.
    % We are first to propose the neural architecture search as automatic search for unsupervised domain adaptation in semantic segmentation. 

    \item We show that a tight integration between network search and domain adaptation is necessary, that it is not enough to simply combine existing NAS and UDA.

    \item We propose a novel evaluation metric for model selection to guide the search controller so that it finds architectures close to those that would result if target domain labels were available.
    % a parameter mapping paradigm  to reduce the pretraining cost and fasten the automatic adaptation process. 
\end{itemize}

%%%We make the following contributions in this work:
%%%% \vspace{-5pt}
%%%\begin{itemize}
%%%\vspace{-6pt}
%%%    \item We approach UDA from a novel perspective, namely \textit{architecture}. We propose AutoAdapt, the first framework to our knowledge to perform automated network search for UDA.
%%%    % We are first to propose the neural architecture search as automatic search for unsupervised domain adaptation in semantic segmentation. 
%%%    \vspace{-10pt}
%%%    \item We show that a tight integration between network search and domain adaptation is necessary, that it is not enough to combine existing NAS and UDA.
%%%    \vspace{-10pt}
%%%    \item We propose a novel evaluation metric for model selection which can guild the search controller to search architectures close to the situation if we could have labels. 
    % a parameter mapping paradigm  to reduce the pretraining cost and fasten the automatic adaptation process. 
%%%\end{itemize}

\section{Related work}
\label{sec:related}

\paragraph{UDA for semantic segmentation} Various UDA techniques have been applied for semantic segmentation. Adversarial training~\cite{tsai2018adaptseg,vu2019advent, tsai2019patch, chen2018road, luo2019clan, vu2019dada, hoffman2018cycada, li2019bdl, Deng_2021_WACV}, has been used where one network predicts the segmentation mask of an image, which can be from either the source or target domain, and another network tries to discriminate which domain the image and mask are from. A second category of methods use self-training to iteratively adapt models using target-set pseudo labels generated by the previous state of the model~\cite{zou2018cbst, li2020ccm, lian2019pycda, zhu2020improving}. Other frameworks to UDA have also been proposed ~\cite{du2019ssf, subhani2020scale-invariant, huang2020contextual, li2020spatial, chang2019structure,yang2020fda, wang2020differential, kim2020texture}. However, all the above methods approach UDA from the image, feature or output-space level. In contrast, out AutoAdapt framework investigates UDA at the architecture level.

%%%\paragraph{UDA in semantic segmentation} Various techniques are applied in UDA for semantic segmentation. Adversarial training\cite{tsai2018adaptseg,vu2019advent, tsai2019patch, chen2018road, luo2019clan, vu2019dada, hoffman2018cycada, li2019bdl}, which generally consists of two networks. One predicts the segmentation mask of the input image from either source or target while the other is the discriminator trying to predict whether the images are from source or target. The second category of methods are build based on self-training. Self-training is to adapt the model with pseudo labels generated for unlabeled target dataset from the previous state of the model\cite{zou2018cbst, zou2018crst, li2020ccm, lian2019pycda}. And there are some other framework have been applied to UDA\cite{du2019ssf, subhani2020scale-invariant, huang2020contextual, li2020spatial, chang2019structure,yang2020fda, wang2020differential, kim2020texture}. All the above UDA methods are developed on either image-level, feature-level or output-space level. We adopt some of the DA methods to help build AutoAdapt, but our contribution is on the architecture-level, where our model architecture is different from all of them.

% These methods are based on the idea of learning image-level or feature-level transformations between the source and the target domains. 

\begin{figure*}[t!]
    \centering
    \includegraphics[width=\linewidth]{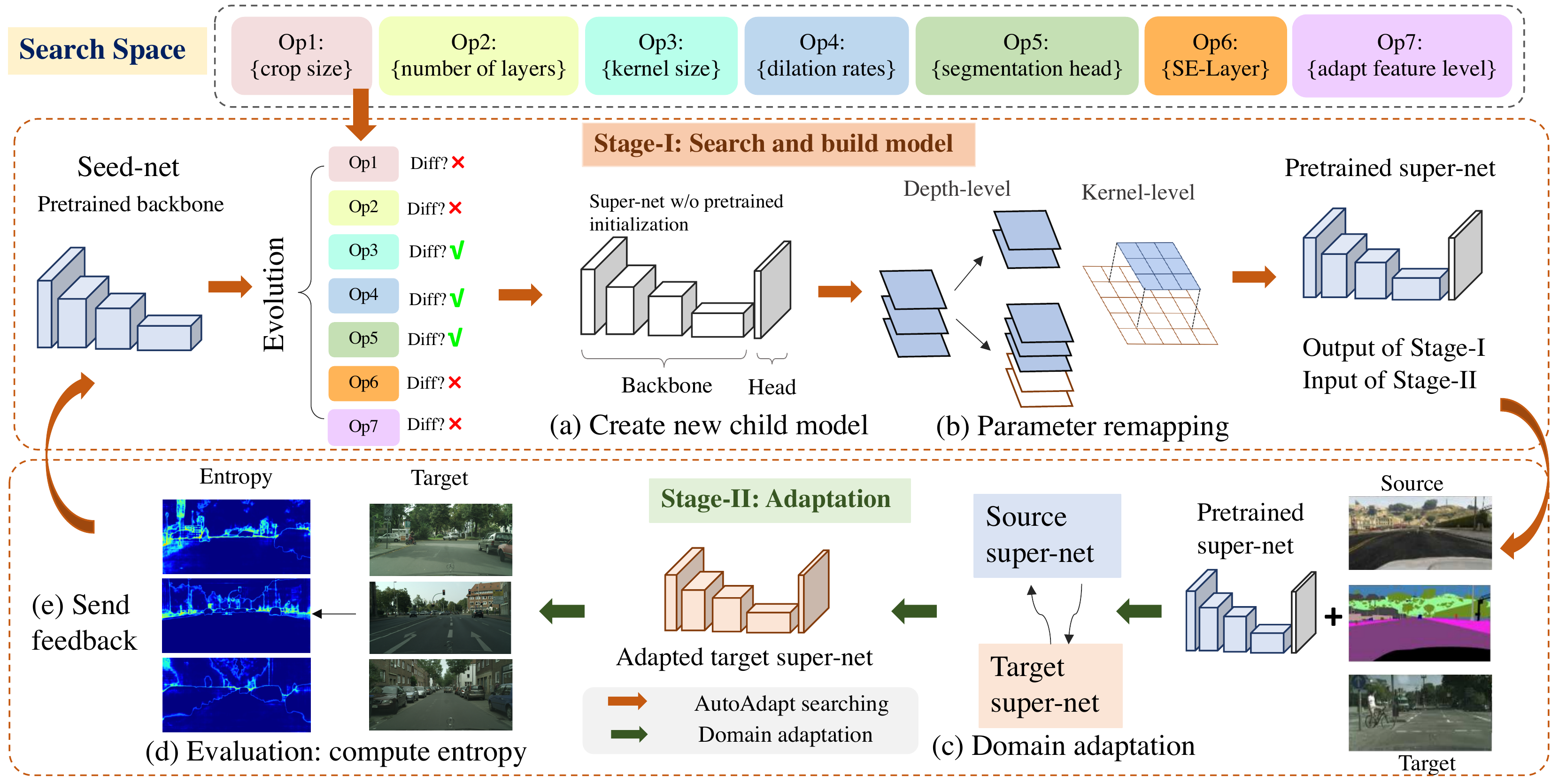}

    \caption{The proposed AutoAdapt. \textbf{Stage-I}: A pretrained backbone (seed-net) is modified by evolutionary algorithm. If the randomly selected $Op_i$ is differentiable then the proposed differentiable mutation is used to create new child model otherwise we follow standard random selection. Once the child model (super-net) is created, parameter remapping is used to migrate the weights from the pretrained seed-net to the super-net. Afterwards, at \textbf{Stage-II}, source images and their annotations as well as the target images are used to perform domain adaptation on the pretrained super-net. Finally, the proposed metric is used to evaluate the adaptation performance, and evolution updates the population to repeat Stage-I again. Details can be found in Section \ref{sec:methods}.  }

    \label{fig:pipeline}
\end{figure*}

\paragraph{AutoML} AutoML has been applied recently to search solutions to problems such as augmentation policy~\cite{lin2019online}. In the context of NAS, numerical techniques have been proposed to compress or find a tiny model architectures. Recently, \cite{liu2018darts} proposed a gradient-based method (DARTS) to make the architecture search differentiable. This makes it faster than other search methods based on reinforcement learning~\cite{zoph2016nas_reinforce} or evolutionary algorithms~\cite{real2019amoebanet,liu2018hierarchical} and reduces the cost from hundreds or thousands of GPU days to just a few GPU. This has increased the scope of problems that NAS has been applied to. For example, DARTS has been applied to search model architectures for generative networks for image synthesis~\cite{gong2019autogan, gao2020nasgan}, for semantic segmentation~\cite{liu2019autodeeplab} and for object detection~\cite{ghiasi2019nas-fpn}, etc. However, to our knowledge, we are the first to search model architectures for unsupervised domain adaptation. %%%%Although, evolutionary algorithm has been applied for some supervised tasks\cite{real2019amoebanet, liu2018hierarchical}, but our proposed evolutionary algorithm overcomes the challenge of lacking ground-truth in this paper.

\section{AutoAdapt: Domain adaptation at the architecture level}
\label{sec:methods}

We now describe our novel AutoAdapt framework for unsupervised domain adaptation in semantic segmentation. We provide background material and identify and analyze the challenges in searching network architectures for UDA in Sec.~\ref{sec:prelim}. We illustrate how we address these challenges by proposing a novel joint evaluation criterion for model selection in Sec.~\ref{sec:evaluation}. Finally, we summarize our proposed AutoAdapt framework in Sec.~\ref{subsec:framework}. Algo.~\ref{algo:autoadpat} provides an overview of AutoAdapt.

%%%In this section, we introduce the proposed AutoAdapt framework for unsupervised domain adaptation in semantic segmentation. We first introduce the background knowledge then identify and analyze the challenges in searching network architectures for UDA in Sec.~\ref{sec:prelim}. We then illustrate how we address these challenges by proposing an novel join evaluation criterion for model selection in Sec. \ref{sec:evaluation}. Finally, we summarize our proposed AutoAdapt framework in Sec.~\ref{subsec:framework}. Algo.~\ref{algo:autoadpat} provides an overview of AutoAdapt.

\subsection{Preliminaries and motivation}
\label{sec:prelim}
NAS searches the space of valid architectures guided by an objective function. If the space of architectures is parameterized by $\alpha$ and the network weights are parameterized by $w$ then the joint architecture and weight search for a specific problem can be formulated as minimizing an objective function $\mathcal{J}(\alpha,w)$ as follows:
%%%Neural architecture search is designed to search the solution of architecture for the objective function. The parameter for building architectures is denoted as $\alpha$, and the parameter for optimizing networks is denoted as $w$. Given an objective function for searching a architecture solution for specific problem, $\mathcal{J}(\alpha,w)$, the optimization for searching is concluded as follow:
\begin{equation}
\label{eq:nas}
    \begin{split}
      & \underset{\alpha}{\text{min}}  \: \mathcal{J}_{search}(\alpha, w^*({\alpha})) \\
       & s.t. \: w^{*} (\alpha) = \underset{w}{\text{argmin}} \:\mathcal{L}_{train}(\alpha, w) .
\end{split}
\end{equation}
This implies a bilevel optimization problem. In order to optimize $\alpha$, we first need to optimize $w$. For most supervised tasks, $\mathcal{J}_{search}$ and $\mathcal{L}_{train}$ can be easily minimized by using a training set to update $w$ and a validation set to update $\alpha$. This is because, the proxy datasets (training and validation) are derived from the same dataset. However, the situation is different for UDA where accuracy can be only computed for the source dataset $(\mathbf{x}^{\mathcal{S}} \in \mathcal{X}^{\mathcal{S}}, \mathbf{y}^{\mathcal{S}} \in  \mathcal{Y}^{\mathcal{S}})$. Due to the lack of labels, the accuracy for target dataset $(\mathbf{x}^{\mathcal{T}} \in \mathcal{X}^{\mathcal{T}})$ cannot be computed and so there is no way to update $\alpha$. 

%%%This implies a bilevel optimization problem. In order to optimize $\alpha$, we first need to optimize $w$. For most supervised tasks, $\mathcal{J}_{search}$ and $\mathcal{L}_{train}$ can be easily achieved by creating a training set for updating $w$ and validation set for updating $\alpha$. This is because, the proxy datasets (train and validate) are derived from the same dataset. However, the situation is different from UDA where accuracy can be only computed for source dataset. Due to lack of label, accuracy of target dataset can't be computed which brings the obstacles to send feedback for updating $\alpha$. 

Focusing on the details of UDA approaches helps explain this problem. We take adversarial learning as an example since it is a common approach to UDA~\cite{tsai2018adaptseg,tsai2019patch, vu2019advent, pan2020intrada}. Since the source dataset has class labels, it can be incorporated into a segmentation loss such as a 2D cross entropy loss. However, the target dataset does not have class labels and so can only be incorporated into an adversarial loss such as a binary cross entropy loss in which the domains become the labels. While the class labels provide semantic information, the domain labels only help the model identify if the input is from source (1) or target (0) in order to align the distributions. We have the following situation
%%%We bring the detailed techniques in UDA in order to further explain the problem. We take adversarial learning as an example which s one of the common approaches for UDA\cite{tsai2018adaptseg,tsai2019patch, vu2019advent, pan2020intrada}. We know that source dataset has class labels,  and it can be optimized by segmentation loss which is a 2D cross entropy loss. While target target can be only optimized by adversarial loss which is a binary cross entropy loss with domain labels. Clas labels can provide semantic information, however, domain label can only help the model to identify if the input is from source (1) or target (0) in order to align distribution.  Details are as follows:
% \vspace{-15pt}

\begin{equation}
\label{eq:source}
     \mathcal{L}_{seg}^{\mathcal{S}}(\mathbf{x}^{\mathcal{S}} , \mathbf{y}^{\mathcal{S}} )={-\mathbf{y}^{\mathcal{S}}\log (\mathbf{p}_{\mathbf{x}^{\mathcal{S}}}})
\end{equation}

\begin{equation}
\label{eq:target}
    \mathcal{L}_{adv}^{\mathcal{T}}(\mathbf{x}^{\mathcal{T}} )=-\log(\mathcal{D}(\mathcal{G}(\mathbf{x}^{\mathcal{T}})))
\end{equation}

\noindent where $\mathcal{S}$ and $\mathcal{T}$ denote the source and target datasets, and the corresponding losses are denoted as $\mathcal{L}^{\mathcal{S}}_{seg}$ and $\mathcal{L}^{\mathcal{T}}_{adv}$ which are the segmentation loss (2D cross-entropy loss with class labels) and the adversarial loss (binary cross entropy loss with domain labels). $\mathcal{G}(\mathbf{x}^{\mathcal{T}})$ is the segmentation network treated as a generator which can operate in the feature, pixel or output space. $\mathcal{D}$ is the discriminator network. Information from the segmentation model (feature, pixel, output space, etc.) is fed to $\mathcal{D}$ to try fool it. $\mathbf{p}_{\mathbf{x}^{\mathcal{S}}}$ denotes the probability output by the segmentation network. The segmentation network is then jointly optimized as follows:
%%%$\mathcal{S}$ and $\mathcal{T}$ denote source and target datasets individually, and the corresponding losses are represented as $\mathcal{L}^{\mathcal{S}}_{seg}$ and $\mathcal{L}^{\mathcal{T}}_{adv}$ which are segmentation loss (2D cross-entropy loss with class labels) and adversarial loss(binary cross entropy loss with domain labels). $\mathcal{G}(\mathbf{x}^{\mathcal{T}})$ can be feature, pixel, or output space. $\mathcal{D}$ is the discriminator network. Information from segmentation model (feature, pixel, output space, etc.) is fed through $\mathcal{D}$ trying to fool it. $\mathbf{p}_{\mathbf{x}^{\mathcal{S}}}$ denotes the probability from the segmentation network. Then the segmentation network will be optimized jointly as follow:

\begin{equation}
\label{eq:seg}
   \underset{\mathcal{G}}{\text{min}} \:  \mathcal{L}_{seg}^{\mathcal{S}}(\mathbf{x}^{\mathcal{S}} , \mathbf{y}^{\mathcal{S}} ) + \lambda  \mathcal{L}_{adv}^{\mathcal{T}}(\mathbf{x}^{\mathcal{T}} )
\end{equation}

Once the segmentation network is updated, the discriminator network is optimized as follows:
\begin{equation}
\label{eq:discriminator}
     \underset{\mathcal{D}}{\text{min}} \:-(1-z)\log (1-\mathcal{D}(\mathbf{p}))+z\log \mathcal{D}(\mathbf{p})
\end{equation}
\noindent where z = 0 if the sample is drawn from the target domain, and z = 1 if it is from the source domain. This two-player game forces the model to generate predictions from the target domain that are close to predictions from the source.

%  For example, . And even for object detection, a more complex objectives including classification loss and bounding box regression loss can be used for the bilevel optimization as well. 

We can see from the above equations that $\mathcal{L}_{adv}^{\mathcal{T}}(\mathbf{x}^{\mathcal{T}} )$ is only an adversarial training loss and so is meaningless for evaluating the model performance on the target domain. The situation is thus different from NAS for supervised problem where $\mathcal{J}_{search}$ is equal to $\mathcal{L}_{val}$ which is the same loss as $\mathcal{L}_{train}$ (training set) but just with a different dataset (validation set). In the supervised NAS problem, $\mathcal{L}_{val}$ reflects how well the model is performing. However, when NAS is used for UDA, Eq.~\ref{eq:seg} and Eq.~\ref{eq:discriminator} form the $\mathcal{L}_{train}$ to update $w$. Since the adversarial loss provides no feedback on how the model performs on the target dataset, $\alpha$, and thus the model architecture, cannot be updated in a meaningful way. We identify this as an optimization gap when applying NAS to UDA.

%%%We can see from the above equations, $ \mathcal{L}_{adv}^{\mathcal{T}}(\mathbf{x}^{\mathcal{T}} )$ is an adversarial training loss which is meaningless for evaluating target domain. The situation is different from that of NAS for supervised application, where $\mathcal{J}_{search}$ equals to $\mathcal{L}_{val}$ which is the same loss as $\mathcal{L}_{train}$ (training set) with a different dataset (validation set). While searching, $\mathcal{L}_{val}$ can reflect how well the model is performing. However, when NAS is performed for UDA with Eq.\ref{eq:nas}, Eq.\ref{eq:seg}  and Eq.\ref{eq:discriminator} are treated as $\mathcal{L}_{train}$ to update $w$. While updating $\alpha$, since the adversarial loss can't provide feedback evaluating target dataset, and thus $\alpha$ can't improve performance on target. We conclude this is an optimization gap for UDA.

\begin{algorithm}[t]
% \SetAlgoLined

\While{not converged}

\textbf{StageI}: Search and build the model.

\begin{algorithmic}[1]
    \STATE Create a child model parameterized by $\alpha$ from search space by search controller.
\end{algorithmic}

\textbf{StageII}: Train the model. 

\begin{algorithmic}[1]
  \STATE Update weights $w$ with domain adaptation loss:

    \STATE Update $\mathcal{G}$: $\underset{\mathcal{G}}{\text{min}} \:  \mathcal{L}_{seg}^{\mathcal{S}}(\mathbf{x}^{\mathcal{S}} , \mathbf{y}^{\mathcal{S}} ) + \lambda  \mathcal{L}_{adv}^{\mathcal{T}}(\mathbf{x}^{\mathcal{T}} )$
     
     \STATE Update $\mathcal{D}: \underset{\mathcal{D}}{\text{min}} \:-(1-z)\log (1-\mathcal{D}(\mathbf{p}))+z\log \mathcal{D}(\mathbf{p})$
   
\end{algorithmic}

% \endwhile
\textbf{Evaluation:} Estimate the performance of $\alpha$ and send it to the controller: 
$\text{MMD}(\mathcal{X}^{\mathcal{S}}, \mathcal{X}^{\mathcal{T}}) + \text{ReEnt}(w^*({\alpha}), \alpha, \mathcal{X}^{\mathcal{T}})$ 

\textbf{{end}}

\textbf{\textit{Derive architecture $\alpha^{*}$ for UDA} and use it for retraining adaptation methods. }

\caption{AutoAdapt}
 \label{algo:autoadpat}
 
%   \vspace{-pt}
\end{algorithm}

\subsection{Bridging the gap without labels}
\label{sec:evaluation}

Then main challenge of the search problem is bridging this optimization gap that results when labels are not available for the target domain. The form of the objective used to guide the search is an open problem. Further, it is clear that the performance of a candidate model on the target dataset can only be approximated. We seek a joint metric to approximate this performance. Specifically, maximum mean discrepancy (MMD) is used for computing the distance between source and target domain and entropy is used for determining the model confidence based on output structure. 

%%%Then main challenge of the search problem is how to bridge the optimization gap where labels are not provided for target domain. Therefore this makes the searching objective an open question, that is only approximation can be achieved for evaluating the performance on target domain. We turn to seek a joint metric to approximate the performance. In details, maximum mean discrepancy (MMD) is used for computing the distance and entropy is used for reporting the model confidence. 

\noindent \textbf{Maximum Mean Discrepancy (MMD)}~\cite{gretton2008mmd} has been widely used as a regularizer for domain adaptation~\cite{saito2018mmd_da,long2017deep,tzeng2014deep}. We propose using it as metric to measure the distance between the source and target domains for evaluating a model's adaptation performance: 

\begin{equation}
    \text{MMD}[\mathcal{S},\mathcal{T}]={\| {\mathbb{E}}_{\mathcal{S}} [\phi{(X^{\mathcal{S}}})}] - \mathbb{E}_{{\mathcal{T}}} [\phi{(X^{\mathcal{T}})] \|}_{\mathcal{H}}
\end{equation}
\noindent where $\phi(.)$ represents the feature space and $\mathcal{H}$ represents a Reproducing Kernel Hilbert Space (RKHS). More details can be found in~\cite{gretton2008mmd}.

If the two domains are drawn from the same distribution, the MMD will be zero. However, this is not possible for domain adaptation where the domains are drawn from different distributions. Hence, MMD should be as small as possible. In other words, in model selection, a smaller MMD may indicate a better model, one that reduces the domain gap. But, this is not enough for UDA in semantic segmentation. Since semantic segmentation is a pixel-wise classification problem, a small distance between the global feature distributions at the image scale does not guarantee good pixel-wise performance. We therefore explore using entropy to also evaluate the structure of the model output.

%%%If the two domains are drawn from the same distribution, the MMD will be zero. However, this is impossible for domain adaptation which are drawn from different distributions. Hence, MMD should be as small as possible. In other words, in model selection, a smaller MMD may indicate a better model which reduces a smaller domain gap. But it is still not enough for UDA in semantic segmentation. Since semantic segmentation is pixel-wise classification problem, a small distance between the global feature distributions at image scale is not guaranteed for pixel-wise performance. We explore to use entropy for evaluating the output structure.

\noindent \textbf{Regional Weighted Entropy (ReEnt)} Entropy has been used in domain adaptation~\cite{vu2019advent, pan2020intrada} beyond feature and output spaces. As a confidence measurement, entropy indicates where a model is struggling. We therefore investigate using entropy to evaluate how candidate architectures perform on the target dataset. 
%%%\noindent \textbf{Regional Weighted Entropy (ReEnt)} Entropy has been adopted for domain adaptation~\cite{vu2019advent, pan2020intrada} besides feature, output space. As a confidence measurement, entropy indicates where a model is struggling. We therefore investigate using entropy to evaluate how candidate architectures perform on a target dataset. 
% We are the first to propose to use it as a metric in neural architecture search. The details of computing entropy are described as follows.
Specifically, given a softmax score map $\mathbf{p}$ for a target image $\mathbf{x}^{\mathcal{T}}$, denoted as $\mathbf{p}_{\mathbf{x}^{\mathcal{T}}}^{(h,w,c)}$ where $h,w,c \in \mathcal{H}, \mathcal{W},\mathcal{C}$ indicate the indices along height, width and class dimensions of the last softmax layer of the model, we compute the entropy as

\begin{equation}
\small
   \text{Ent}_{\textbf{x}^{\mathcal{T}}}^{(h,w)} =  \frac{-1}{\log (\mathcal{C})} \sum_{c=1}^{\mathcal{C}}{\mathbf{p}_{\mathbf{x}^{\mathcal{T}}}^{(h,w,c)}\log \mathbf{p}_{\mathbf{x}^{\mathcal{T}}}^{(h,w,c)}}.
\end{equation}
Here, $\mathcal{C}$ indicates the number of semantic classes. $ \text{Ent}_{\mathbf{x}^{\mathcal{T}}}^{(h,w)}$ is a 2D matrix, and in order to produce a consistent metric, we average the entropy over all pixel locations as follows: 

\begin{equation}
    \text{Ent}_{\mathbf{x}^{\mathcal{T}}}= \frac{1}{\mathcal{H}\mathcal{W}} \sum_{h=1}^{\mathcal{H}} \sum_{w=1}^{\mathcal{W}}  \text{Ent}_{\mathbf{x}^{\mathcal{T}}}^{(h,w)} 
\end{equation}
  % Table generated by Excel2LaTeX from sheet 'backbone'
\begin{table*}[!ht]
%   \centering
  \resizebox{1.0\linewidth}{!}{%
    \begin{tabular}{l|c|c|ccc|cc|c}
    \toprule
    \multicolumn{1}{c|}{\multirow{2}[1]{*}{Segmentation Model}} & \multirow{2}[1]{*}{Backbone} & \multirow{2}[1]{*}{Segmentation Head} &\multicolumn{3}{c|}{Domain Adaptation (mIoU)}   &\multirow{2}[1]{*}{$\#$ Params}
    &\multirow{2}[1]{*}{$\#$ MACs}
    &\multirow{2}[1]{*}{Oracle} \\
          &   & &{AdvEnt} & {IntraDA} & {CCM} & & & \\
          \hline
          
    DeepLabV2\cite{chen2017deeplabv2} & Res101 & ASPP &43.8  & 45.8  & 49.9  & 44.6M & 380.5B &65.1 \\
   
    DANet \cite{fu2019danet}& Res101   & Attention & 39.8    & 42.7  &  47.5& 57.7M &490.1B  & 77.6 \\

    DeepLabV3+\cite{chen2018deeplabv3plus} & Res101 &   ASPP and Low-level    & 42.5  & 45.4 &  47.1&68.6M&627.0B&77.8 \\
    DeeplabV2-S$^*$& SENet\cite{hu2018senet}& ASPP  & 41.3  & 43.9  & 45.7 & 49.3M&381.6B  & $-$\\
    \hline
     \hline
    Deeplabv2-D\cite{liu2018darts} & Searched   & ASPP & 29.8  &  32.1  &  36.8 &5.0M&6.2B&$-$ \\
    AutoDeeplab\cite{liu2019autodeeplab} & Searched     & Searched & 37.6  &  39.4  & 42.9& 10.2M &33.2B &79.7 \\
    \hline
    \hline
    Random search  & NAS   & NAS& 37.8 &  38.5  &39.4 & 45.5M & 381.2B &$-$ \\
    DARTS-1$^*$  & NAS   & ASPP& 24.8 &  27.1 & 28.8  & 5.0M & 6.2B &$-$ \\
    DARTS-2$^*$  & NAS   & ASPP  &   28.4 & 33.9& 39.8& 5.0M& 6.2B&$-$ \\
    PC-DARTS$^*$ & NAS & ASPP & 35.7 & 39.8& 43.1& 8.0M& 8.7B&$-$ \\
    AutoDeeplab$^*$ & NAS   & NAS&     32.8 &35.1& 37.5& 10.2M & 33.2B &$-$ \\
    \hline
    \hline
    AutoAdapt (MMD)  & NAS   & NAS&    44.2   & 45.9  & 49.8 & 44.7M & 380.1B& $-$ \\
    AutoAdapt (Ent.)  & NAS   & NAS & 44.3  & 46.0 & 49.9 & 44.8M  &380.5B &$-$ \\
    AutoAdapt (ReEnt) & NAS   & NAS&   44.6  & 46.5 & 49.9 & 44.8M  &380.5B &$-$ \\ 
    AutoAdapt (Joint)  & NAS   & NAS& \textbf{45.3}  & \textbf{47.2}    & \textbf{50.2}  &44.9M&380.5B & $-$ \\
    \bottomrule
        \end{tabular}%
    }
      \caption{Comparisons between automatic search and hand-crafted architectures for UDA. mIoU is reported as the evaluation metric. DeepLabV2-S denotes our implementation with backbone SENet, and DeeplabV2-D denotes our implementation with backbone DARTS which is searched on CIFAR and then trained on ImageNet. * denotes our implementation. The difference between searched and NAS is : Searched: searched by other application and train for UDA; NAS: perform searching for UDA and train for UDA.}
  \label{tab:overall_comp}%
\end{table*}%

However, average entropy does not achieve our goal of evaluating the structure of the output. We therefore propose an improved \textbf{Re}gional weighted \textbf{Ent}ropy \textbf{(ReEnt)} to measure adaptation performance with respect to output structure. We first partition the input image into regions $\mathcal{R}$ based on the appearance similarity computed using the low-level features. We then compute the entropy value for each region as follows:
%%%However, the average entropy can't meet our expectation which should describe the output structure. We hence propose an improved \textbf{Re}gional weighted \textbf{Ent}ropy \textbf{(ReEnt)} to depict the adaptation results regarding output structure. We first split the input image into regions $\mathcal{R}$ based on the appearance similarity computed by the low-level feature. We then compute the entropy value for each region as follow:
% \textbf{\color{red} check the correctness}

\begin{equation}
    \text{ReEnt}_{\textbf{x}^{\mathcal{T}}}^{(r, c)} = \beta \frac{1}{\log (r)} \sum_{h^{\prime},w^{\prime}}^{r}  \mathbf{p}_{\textbf{x}^{\mathcal{T}}}^{(h^{\prime},w^{\prime}, c)} \log \mathbf{p}_{\textbf{x}^{\mathcal{T}}}^{(h^{\prime},w^{\prime}, c)} 
\end{equation}
where $h^{\prime},w^{\prime}$ represent the location of the region $r$. Since the areas of the regions vary, $\beta$ is used to weight their importance as measured by their proportion of the total area:

\begin{equation}
    \beta = \frac{Area^{r}}{ \sum_{r^{\prime}=1}^{\mathcal{R}} Area^{r^{\prime}}}
\end{equation}

This is different from standard entropy which is used to characterize the probability distribution along class dimensions. Our goal instead is to determine if a model produces smooth predictions within regions that have similar appearance. If the probability is evenly distributed (smooth prediction) within a region, then the entropy will be large entropy\cite{shannon1948mathematical}. So that all the metrics are consistent (smaller is better), we take the opposite value. Finally we derive the regional weighted entropy for the whole image as:
%%%Different from the original entropy which is used to measure the probability distribution along class dimension.  Our goal is to measure if the model is providing smooth prediction within the region where similar appearance is shown. If the probability is evenly distributed (smooth prediction) within the region, then it will result in a large entropy. In order to keep all the metric consistent (smaller the better), we need to take the opposite value in which . Finally we can derive the regional weighted entropy for the whole image as follow:
\begin{equation}
    \text{ReEnt}_{\mathbf{x}^{\mathcal{T}}}=\frac{1}{\mathcal{C}}\sum_{c=1}^{\mathcal{C}} \sum_{r}^{\mathcal{R}} \text{ReEnt}_{\textbf{x}^{\mathcal{T}}}^{(r, c)}
\end{equation}

\noindent \textbf{Joint Metric for Model Selection} Our metric for evaluating the performance of UDA which is also used for model selection consists of two parts, one to evaluate domain alignment and another to evaluate output structure. In order to also limit the computational cost of the searched segmentation models, we add a computation complexity (MACs) constraint as a regularization term in the objective function. Our search goal thus becomes minimizing the objective function as follows: 
%%%\noindent \textbf{Joint Metric for Model Selection} Therefore the whole metric for model selection consists of two parts, evaluating domain alignment and output structure. In order to control the computational cost of the searched segmentation model at the same time, we add the computation complexity (MACs) constraint as a regularization term in the objective function. Our search goal thus becomes  minimizing the objective function as follow: 
% \vspace{-10pt}
\begin{equation}
\label{eq:objective}
    \begin{split}
       \underset{\alpha}{\text{min}}  \: & \text{MMD}(\mathcal{X}^{\mathcal{S}}, \mathcal{X}^{\mathcal{T}}) + \text{ReEnt}(w^*({\alpha}), \alpha, \mathcal{X}^{\mathcal{T}}) + \lambda \text{MACs}(\alpha) \\
      & s.t. \: w^{*} (\alpha)  = \underset{w}{\text{argmin}} \:\mathcal{L}_{DA}(\alpha, w, \mathcal{X}^{\mathcal{S}}, \mathcal{X}^{\mathcal{T}}) .
\end{split}
% \vspace{-0pt}
\end{equation}

It is noted that, the proposed evaluation metric can not be served as the training objective. Since the proposed ReEnt is not differentiable resulting in not available for back propagation while training.

\subsection{Framework}

\label{subsec:framework}
We now summarize the proposed AutoAdapt framework for unsupervised domain adaptation. As is done in most NAS techniques, we create a proxy task to perform NAS. Once the optimal architecture is determined, the model is retrained using the entire dataset.

As shown in Algo.~\ref{algo:autoadpat}, AutoAdapt has two stages. \textbf{Stage I: Search and build model} Given a specified search space and a pretrained backbone (seed-net), a child model (super-net) will be created by the search controller (which could be based on reinforcement learning, evolutionary algorithms, gradient descent, etc.).  \textbf{Stage II: Domain adaptation} Given the child model, our goal is to update the model weights by adapting from the source to the target, that is to compute $w^{*}(\alpha)$. Note that any domain adaptation method can be used to accomplish this. We then evaluate the adapted model using the proposed joint metric $\text{MMD}(\mathcal{X}^{\mathcal{S}}, \mathcal{X}^{\mathcal{T}}) + \text{ReEnt}(w^*({\alpha}), \alpha, \mathcal{X}^{\mathcal{T}})$, and, through the optimization in Eq.\ref{eq:objective}, provide feedback to the controller to generate the next novel architecture. Through such interactions, the final optimal solution will be derived from the minimum evaluation score. To improve efficiency, the search is based on a proxy task, that is to find $\alpha$ using a small number of  training iterations and with a simple UDA method (AdvEnt\cite{vu2019advent}). Finally, the discovered architecture $\alpha^*$ is retrained with the full datasets and with a sufficient number of iterations. Note that we can consider different UDA methods during retraining. In the experiments below, we investigate AdvEnt~\cite{vu2019advent}, IntraDA~\cite{pan2020intrada} and CCM~\cite{li2020ccm}.

 \begin{figure*}[!ht]
    \centering
    \includegraphics[width=\linewidth]{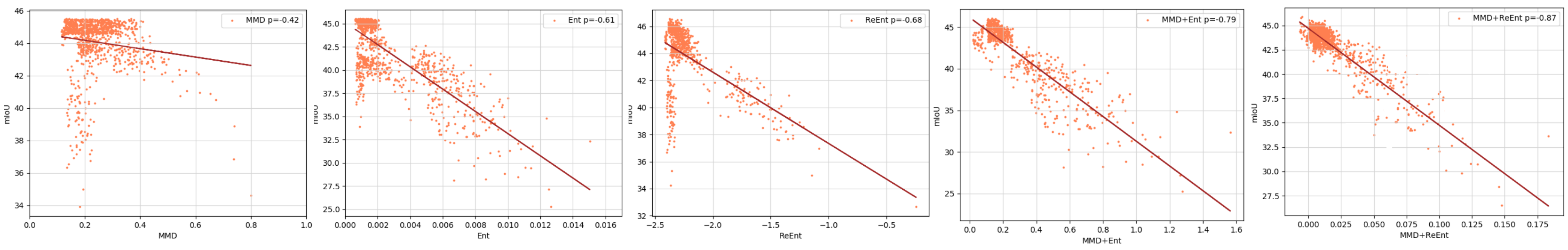}

    \caption{Correlation between entropy and mIoU. Entropy is highly (inversely) correlated with mIoU as indicated by Spearman's rank correlation (p) for all metrics used in AutoAdapt. All experiments are performed on GTA5$\rightarrow$Cityscapes. (Zoom in for details.)}

    \label{fig:ablation}
\end{figure*}

% Table generated by Excel2LaTeX from sheet 'table'
\begin{table*}[t!]
\small
% \begin{subtable}[t][\textwidth]
% \caption{Experimental results on GTA5$\longrightarrow$Cityscapes}
% \centering
\resizebox{1.0\textwidth}{!}{%
    \begin{tabular}{l|p{0.35cm}p{0.35cm}p{0.35cm}p{0.35cm}p{0.35cm}p{0.35cm}p{0.35cm}p{0.35cm}p{0.35cm}p{0.35cm}p{0.35cm}p{0.35cm}p{0.35cm}p{0.35cm}p{0.35cm}p{0.35cm}p{0.2cm}p{0.3cm}p{0.4cm}|c}
    \toprule
          &  \rotatebox{90}{road} & \rotatebox{90}{side.} & \rotatebox{90}{build.} & \rotatebox{90}{wall} & \rotatebox{90}{fence} & \rotatebox{90}{pole} & \rotatebox{90}{light} & \rotatebox{90}{sign} & 
          \rotatebox{90}{veg} & \rotatebox{90}{terr.} & \rotatebox{90}{sky} & \rotatebox{90}{pers.} & \rotatebox{90}{rider} & \rotatebox{90}{car} & \rotatebox{90}{truck} & \rotatebox{90}{bus} & \rotatebox{90}{train} & \rotatebox{90}{mbike} & \rotatebox{90}{bike} & 
          {mIoU} \\
          \hline
    % Table generated by Excel2LaTeX from sheet 'table'
    Source Only    & 60.6  & 17.4  & 73.9  & 17.6  & 20.6  & 21.9  & 31.7  & 15.3  & 79.8  & 18.1  & 71.1  & 55.2  & 22.8  & 68.1  & 32.3  & 13.8  & 3.4   & \textbf{34.1}  & 21.2  & 35.7 \\
    % \hline
     CRST\cite{zou2018cbst}      & 91.0  & 55.4  & 80.0  & 33.7  & 21.4  & 37.3  & 32.9  & 24.5  & 85.0  & 34.1  & 80.8  & 57.7  & 24.6  & 84.1  & 27.8  & 30.1  & 26.9  & 26.0  & 42.3  & 47.1 \\
        PLCA\cite{kang2020cycleDA}    & 84.0 & 30.4  & 82.4  & 35.3  & 24.8 & 32.2 & \textbf{36.8}  & 24.5 & 85.5  & 37.2  & 78.6  & \textbf{66.9}  & 32.8  & 85.5  & 40.4 & 48.0  & 8.8   & 29.8 & 41.8  & 47.7 \\
      BDL \cite{li2019bdl}     & 91.0  & 44.7  & 84.8  & 34.6  & 27.6  & 30.2  & 36.0  & 36.0  & 85.0  & 43.6  & 83.0  & 58.6  & 31.6  & 83.3  & 35.3  & 49.7  & 3.3   & 28.8  & 35.6  & 48.6 \\
     
    CAG \cite{zhang2019cag}      & 90.4  & 51.6  & 83.8  & 34.2  & \textbf{27.8}  & \textbf{38.4}  & 25.3  & \textbf{48.4}  & 85.4  & 38.2  & 78.1  & 58.6  & \textbf{34.6}  & 84.7  & 21.9  & 42.7  & \textbf{41.1}  & 29.3  & 37.2  & 50.1 \\

        \hline
      \hline
    AdvEnt\cite{vu2019advent}    & 89.9  & 36.5  & 81.6  & 29.2  & 25.2  & 28.5  & 32.3  & 22.4  & 83.9  & 34.0  & 77.1  & 57.4  & 27.9  & 83.7  & 29.4  & 39.1  & 1.5   & 28.4  & 23.3  & 43.8 \\
    Ours(Auto-AdvEnt)   & 90.8  & 38.6  & 82.4  & 29.9  & 24.9  & 26.7  & 34.9  & 22.0  & 81.9  & 33.3  & 77.3  & 58.9  & 28.8  & 85.5  & 38.9  & 49.8  & 0.0   & 23.9  & 31.9  & \textbf{45.3} \\
    \hline
    %   \hline

    IntraDA\cite{pan2020intrada}  & 90.6  & 37.1  & 82.6  & 30.1  & 19.1  & 29.5  & 32.4  & 20.6  & \textbf{85.7}  & 40.5  & 79.7  & 58.7  & 31.1  & \textbf{86.3}  & 31.5  & 48.3  & 0.0   & 30.2  & 35.8  & 45.8 \\
        Ours (Auto-IntraDA)  & 91.6  & 41.8  & 83.5  & 30.7  & 25.4  & 28.9  & 35.4  & 23.1  & 85.2  & \textbf{42.5}  & 81.2  & 60.3  & 29.5  & 85.3  & 31.7  & 49.3  & 0.0   & 30.5  & 40.6  & \textbf{47.2} \\
        % \hline
    \hline
    CCM \cite{li2020ccm}      & \textbf{93.5}  & \textbf{57.6}  & 84.6  & 39.3  & 24.1  & 25.2  & 35.0  & 17.3  & 85.0  & 40.6  & \textbf{86.5}  & 58.7  & 28.7  & 85.8  & \textbf{49.0}  & \textbf{56.4}  & 5.4   & 31.9  & 43.2  & 49.9 \\
    
    Ours (Auto-CCM) &  93.3  & 57.2  & 84.6  & \textbf{39.5}  & 26.1  & 28.4  & 34.9  & 20.7  & 84.7  & 40.3  & 86.2  & 60.7  & 29.3  & 85.4  & 48.5  & 55.2  & 6.1   & 30.1  & \textbf{43.4}  & \textbf{50.2} \\
    \bottomrule
    \end{tabular}%
    }
    \caption{Experimental results of GTA5$\rightarrow$Cityscapes. We show results of Auto-AdvEnt, Auto-IntraDA, Auto-CCM. }
      \label{tab:sota-gta}
\end{table*}%

\section{Search strategy}
\label{sec:strategy}
% \vspace{-7pt}
Now that we have described a general framework for search optimization, we can investigate it with a specific search strategy. Note that any search algorithm can be used as the search controller. We find search based on evolutionary algorithms is easily incorporated into our framework and has an acceptable search speed. In this section, we first describe the specific search space we design for UDA. We then present our evolutionary algorithm for searching.
%%%Now that we have established the framework for search optmization, we start to design the algorithm with a specific search strategy. Technically, any search algorithm can be used as search controller. We find evolutionary algorithm is easy to be combined with an acceptable search speed. In this section, we first describe the search space specifically designed for UDA. We then present our evolutionary algorithm. \vspace{-5pt}
\subsection{Search space}

\label{sec:search_space}
The NAS search space defines a set of basic network operations as well as ways to combine these operations to construct valid network architectures.
Here, we design our search space that is unique for domain adaptation. We describe our search space in three aspects: training configurations, segmentation model and adaptation configurations. The operations used in each aspect are listed below. More details can be found in the supplementary materials.
\begin{itemize}

    \item Training configuration: We consider a set of crop sizes (Op1) by multiplying the original image dimensions by reduction ratios of $\{$1/8, 1/4, 1/2 and 3/4$\}$. Once the crop size is determined, random cropping is used to generate the input image.

    \item Segmentation model: We consider both the network backbone and the segmentation head. For the backbone, we design the search space as the number of layers (Op2), the kernel size (Op3), and the dilation rates (Op4). We sample kernel sizes from $\{$$3\times3$, $5\times5$ and $7\times7$$\}$, and dilation rates from $\{$$1$, $2$ and $4$$\}$. We also consider incorporating an SE-Layer~\cite{hu2018senet} (Op6) into the backbone. For the segmentation head (Op5), we choose from $\{$ASPP \cite{chen2017deeplabv2}, spatial attention \cite{fu2019danet} and low-level features 
    \item Adaptation configuration: Multi-level adaptation is commonly used in UDA approaches~\cite{tsai2018adaptseg,vu2019advent} by constructing an auxiliary loss using features from the early layers of the network.
    % In detail, the lower-level feature (the levels except the last level) is used to build an auxiliary loss to help adaptation. 
    We also adopt this and consider features from different layers in our search space.
    % e.g. level-1, level-2, and level-3 in ResNet-101.

\end{itemize}

\subsection{Evolutionary algorithm}

\label{sec:algorithm}
% \vspace{-10pt}
\noindent \textbf{Initialization} 
% The evolutionary method we used is summarized in Algorithm \ref{algo:ea}.
We first randomly sample $100$ models from the search space. Given a pretrained backbone, this sampling is accomplished by randomly changing one part of the backbone, selecting a different training configuration, or selecting a different adaptation method. We train and evaluate these models and pick the top $20$ based on our joint criterion to initialize the \textit{population}.

% \vspace{-10pt}
\noindent \textbf{Evolution} 
After initialization, the evolutionary algorithm is used to improve the population over $M$ cycles. Tournament selection \cite{goldberg1991tournament} is used to update and generate new populations. At each cycle, $25\%$ of the models are randomly sampled as the tournament. The model with the lowest entropy is selected as a \textit{parent}. A new model, called a \textit{child}, is constructed from this parent by modifying one part of the model randomly. This process is called a \textit{mutation}. 
% The mutation could be a change in the number of layers, kernel size, etc. as described in section \ref{sec:search_space}. 
The child model is then trained, evaluated and added back to the population. More details of our evolutionary algorithm can be found in the supplementary material.

 \begin{figure*}[!ht]
    \centering
    \includegraphics[width=\linewidth]{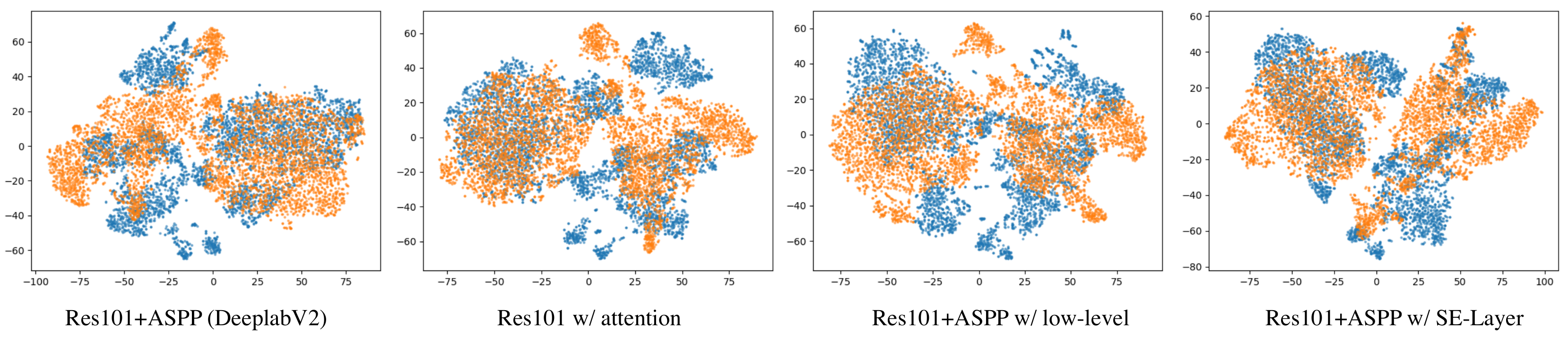}

    \caption{TSNE embeddings. Blue: source domain, orange: target domain (Zoom in for details.)}

    \label{fig:tsne}
\end{figure*}

% Table generated by Excel2LaTeX from sheet 'table'
\begin{table*}[h!]
\small
% \begin{subtable}[t][\textwidth]
% \caption{Experimental results on GTA5$\longrightarrow$Cityscapes}
% \centering
\resizebox{1.0\textwidth}{!}{%
    \begin{tabular}{{l|c|p{0.35cm}p{0.35cm}p{0.35cm}p{0.35cm}p{0.35cm}p{0.35cm}p{0.35cm}p{0.35cm}p{0.35cm}p{0.35cm}p{0.35cm}p{0.35cm}p{0.35cm}p{0.35cm}p{0.45cm}|c|c}}
    \toprule
         &  \rotatebox{90}{road} & \rotatebox{90}{side.} & \rotatebox{90}{build.} & \rotatebox{90}{wall*} & \rotatebox{90}{fence*} & \rotatebox{90}{pole*} & \rotatebox{90}{light} & \rotatebox{90}{sign} & 
          \rotatebox{90}{veg} &
          \rotatebox{90}{sky} & \rotatebox{90}{pers.} & \rotatebox{90}{rider} & \rotatebox{90}{car} &
          \rotatebox{90}{bus} &  \rotatebox{90}{mbike} & \rotatebox{90}{bike} & 
          {mIoU} & mIoU* \\
          \hline
    % Table generated by Excel2LaTeX from sheet 'table'
    Source Only   & 47.1  & 23.3  & 75.6  & 7.1   & 0.1   & 23.9  & 5.1   & 9.2   & 74    & 73.5  & 51.1  & 20.9  & 39.1  & 17.7  & 18.4  & 34    & 34.5  & 40.1 \\
    % \hline
    %  PLCA\cite{kang2020cycleDA} & -    & 82.6  & 29.0  & 81.0  & 11.2   & 0.2   & 33.6  & 24.9   & 18.3   &82.8  &82.3  & 62.1  & 26.5 & 85.6  & 48.9  & 26.8 & 52.2  & 46.8  & 54.0 \\
        MaxSquare \cite{chen2019maxsquare}  &   82.9& 40.7& 80.3 &10.2 &0.8 &25.8 &12.8 &18.2 &82.5& 82.2& 53.1 &18.0 &79.0 &31.4 &10.4 &35.6 &41.4& 48.2 \\
      BDL \cite{li2019bdl}      & \textbf{86.0}  & \textbf{46.7}  & 80.3  & - & - & - & 14.1  & 11.6  & 79.2  & 81.3  & 54.1  & 27.9  & 73.7  & 42.2  & 25.7  & 45.3  & - & 51.4 \\
    CAG \cite{zhang2019cag}     & 84.7  & 40.8  & \textbf{81.7}  & 7.8   & 0.0   & \textbf{35.1}  & 13.3  & \textbf{22.7}  & \textbf{84.5}  & 77.6  & \textbf{64.2}  & 27.8  & \textbf{80.9}  & 19.7  & 22.7  & 48.3  & 44.5  & - \\

    \hline
    \hline
    AdvEnt \cite{vu2019advent}   & 85.6  & 42.2  & 79.7  & - & - & - & 5.4   & 8.1   & 80.4  & 84.1  & 57.9  & 23.8  & 73.3  & 36.4  & 14.2  & 33.0  &  -  & 48.0 \\
        Ours(Auto-AdvEnt)    & 84.8  & 41.9  & 80.3  & 9.5   & 0.3   & 26.8  & 10.2  & 10.8  & 81.2  & \textbf{84.2}  & 59.9  & 25.1  & 73.0  & 39.4  & 13.5  & 35.8  & 42.3  & \textbf{49.2} \\
        \hline
        % \hline
    IntraDA\cite{pan2020intrada}     & 84.3  & 37.7  & 79.5  & 5.3   & \textbf{0.4}   & 24.9  & 9.2   & 8.4   & 80.0  & 84.1  & 57.2  & 23.0  & 78.0  & 38.1  & 20.3  & 36.5  & 41.7  & 48.9 \\
            Ours(Auto-IntraDA) & 84.0  & 40.8  & 79.9  & 7.3   & 0.2   & 25.2  & 12.3  & 11.3  & 81.3  & 82.3  & 59.4  & 26.0  & 79.7  & 40.4  & 21.2  & 36.9  & \textbf{43.0}  & \textbf{50.4} \\
        \hline
        % \hline
  
    % \hline
    % \hline
    CCM  \cite{li2020ccm}     & 79.6  & 36.4  & 80.6  & \textbf{13.3}  & 0.3   & 25.5  & \textbf{22.4}  & 14.9  & 81.8  & 77.4  & 56.8  & 25.9  & 80.7  & \textbf{45.3}  & 29.9  & \textbf{52.0}  & 45.2  & 52.9 \\
    Ours(Auto-CCM)  & 83.6  & 39.1  & 81.6  & 8.5   & 0.3   & 27.4  & 21.7  & 13.2  & 81.9  & 78.0  & 57.9  & \textbf{29.5}  & 79.6  & 43.2  & \textbf{31.8}  & 51.9  & \textbf{45.6}  & \textbf{53.3} \\
    \bottomrule
    \end{tabular}%
    }
%  \end{subtable}
    \caption{Experimental results of SYNTHIA$\rightarrow$Cityscapes. We show results of our approach using Auto-AdvEnt, Auto-IntraDA, and Auto-CCM.  mIoU* denotes the mIoU over per class IoU without wall*, fence*, and pole*.}
      \label{tab:sota-synthia}
\end{table*}%

%  \vspace{-5pt}
\section{Experiments}
\label{sec:experiments}
\subsection{Datasets}

\noindent\textbf{GTA5}: The synthetic dataset GTA5 \cite{richter2016gta5} contains 24,966 synthetic images with a resolution of 1,914$\times$1,052 and corresponding ground-truth annotations. These synthetic images are collected from a video game based on the urban scenery of Los Angeles city. The ground-truth annotations are generated automatically and contain 33 categories. For training, we consider the 19 categories in common with the Cityscapes dataset \cite{cordts2016citys}, similar to previous work. \textbf{SYNTHIA}: SYNTHIA-RAND-CITYSCAPES \cite{ros2016synthia} is another synthetic dataset which contains 9,400 fully annotated RGB images. During training, we consider the 16 categories in common with the Cityscapes dataset. During evaluation, 16- and 13-class subsets are used to evaluate the performance. \textbf{Cityscapes}: A real world dataset, Cityscapes \cite{cordts2016citys} contains 3,975 images with fine segmentation annotations. We use 2,975 images for training and 500 images from the evaluation set to evaluate the performance of our models.

\subsection{Implementation details}

% \textbf{TODO:Not sure if this is still necessary}
% We use 16 V100 GPUs to perform our experiments. 
We use the PyTorch deep learning framework to AutoAdapt. For AutoAdapt Stage-I, we first randomly generate 100 models with ResNet-101 which are are then trained for 20,000 iterations. We the pick the top 20 models as our population and start evolving them through mutation. Once a child model is created, parameter remapping~\cite{fang2020fna} is used to transfer the weights from the pretrained ResNet-101. The model is now ready for to be trained for domain adaptation (Stage-II). We split the target dataset into two folds, 85\% for domain adaptation and 25\% for performance evaluation. We stop the searching early if the best model in the effective population remains the same for over 25 cycles. Once the architecture search is finished, we retrain the model with the full dataset and sufficient training iterations. Three UDA methods AdvEnt~\cite{vu2019advent}, IntraDA~\cite{pan2020intrada} and CCM~\cite{li2020ccm} are used for retraining and thus we term our final results Auto-AdvEnt, Auto-IntraDA and Auto-CCM.  More details can be found in the supplementary materials.

%%%The PyTorch deep learning framework is used to implement our framework.  For AutoAdapt Stage-I, we first randomly generate 100 models with ResNet-101 which are trained over 20,000 iterations. Then we pick the top 20 models as our population. We then start the evolution with mutation. Once the child model is created, parameter remapping\cite{fang2020fna} is used to transfer the weights from pretrained ResNet-101. Afterwards, the model will be ready for training domain adaptation (Stage-II). We split the target dataset into two folds, 85\% for domain adaptation, 25\% for evaluating its performance. We early stop searching if the best model in the effective population remains the same for over 25 cycles. Once the searching is finished, we can do the retraining with full dataset and full training iterations. Three UDA methods AdvEnt\cite{vu2019advent}, IntraDA\cite{pan2020intrada}, and CCM\cite{li2020ccm} are used for retraining and thus termed Auto-AdvEnt, Auto-IntraDA and Auto-CCM.  More details can be found in supplementary materials.

\section{Experimental results}

\label{sec:results}
\subsection{Effectiveness of AutoAdapt}
\label{sec:baselines}

In this section, we perform comparisons to answer the question ``Is automated model search really necessary and effective for UDA?'' That is, how does automated model search for UDA compare with selecting hand-crafted architectures and then applying UDA. We simulate manual design by humans by selecting various advanced backbones and segmentation networks and follow this with different UDA methods. Details of the results and methods are shown in Tab.~\ref{tab:overall_comp}. Adaptation is conducted from GTA5$\rightarrow$Cityscapes. We report mIoU on the Cityscape validation set.

%%%In this section, we perform comparison studies to answer the question ``Is automated model search really necessary and effective for UDA?'' That is, how does automated model search for UDA compare with selecting hand-crafted architectures and then applying UDA. We simulate manual design by humans by selecting various advanced backbones and segmentation networks and consider various UDA methods. Details of the results and methods are shown in Tab.~\ref{tab:overall_comp}. Adaptation is conducted from GTA5$\longrightarrow$Cityscapes. We report mIoU on the validation set in Cistyscapes.

 \noindent\textbf{1) Comparisons to hand-crafted methods} we replace the commonly used semantic segmentation network DeeplabV2~\cite{chen2017deeplabv2} with the state-of-the-art networks DeeplabV3+~\cite{chen2018deeplabv3plus} and DANet~\cite{fu2019danet} in a UDA framework. For fair comparison, the backbone remains the same for all three segmentation models. Rows 1-3 show the results of using advanced segmentation networks with various UDA methods. We can see DANet and DeeplabV3 outperform DeeplabV2 in the non-domain adaptation case (oracle=training using the target dataset) by large margins of over 10\%. But, when there is domain adaptation, they perform worse than DeeplabV2, often with over 2\% lower mIoU.  We also investigate how state-of-the-art backbone affects UDA. We replace the backbone of DeeplabV2 which is a ResNet-101 network with SENet~\cite{hu2018senet} and call this DeeplavV2-S. We realize there are other advanced backbones like Xception~\cite{chollet2017xception} that could be considered. But, for fair comparison, we only consider SE-Net\cite{hu2018senet} as it is included in our search space. Row 4 shows the results for DeeplabV2-S. We observe that an advanced backbone does not improve performance as it achieves only 42.3\%, 43.9\% and 45.7\% mIoU when used with AdvEnt, IntraDA and CCM which is lower than the DeeplabV2-ResNet101 results. We conclude therefore that manually designing segmentation networks (backbone or head) for non-domain adaptation scenarios does not necessarily lead to improved UDA results. This raises the question ``what kinds of modules/architecture lead to improvement for UDA?''. We answer this question by providing architecture-level analysis and conclusions in Sec.~\ref{sec:arch_level_analysis}.
 
 \noindent\textbf{2) Comparisons to searched methods} We select the NAS discovered backbone DARTS~\cite{liu2018darts} with an ASPP (DeeplabV2-D) in DeeplabV2 and the NAS discovered segmentation network AutoDeeplab~\cite{liu2019autodeeplab} to separately replace the segmentation networks for different UDA methods. Rows 5 and 6 show these architectures perform poorly for UDA. We conclude that replacing the network with architectures discovered using existing NAS techniques is not effective for UDA. This is because these architectures are discovered in supervised manner which does not take into account the transferability/generalization of the models.

 \noindent\textbf{3) Comparisons to NAS methods} We modify the search loss function with a UDA loss in DARTS~\cite{liu2018darts}, PC-DARTS~\cite{xu2020pcdarts} and AutoDeeplab~\cite{liu2019autodeeplab} to see if existing NAS methods can find effective architectures for UDA. Rows 8-11 show the performance of the modified NAS methods with a UDA loss. Replacing the search loss function does not provide a reasonable search space for UDA due to the optimization gap. The difference between DARTS-1 and DARTS-2 is the input size which results in different network depths due to GPU memory constraints. PC-DARTS is seen to perform better because it allows us to set a larger number of network depths. Finally, we see from the last row in Tab.~\ref{tab:overall_comp} that our proposed AutoAdapt approach which performs architecture search specifically for UDA provides the best performance for all three domain adaptation methods. 
 
%  We find it interesting that all the hand-crafted networks perform worse than the DeepLabV2 baseline in the UDA case. This further motivates the need to approach UDA at the architecture level as is done by the proposed AutoAdapt.

 \noindent\textbf{4) Comparison to state-of-the-art} We compare our method to state-of-the-art domain adaptation methods on two unsupervised domain adaptation tasks: GTA5$\rightarrow$ Cityscapes and SYNTHIA$\rightarrow$Cityscapes. The results are presented in Tabs.~\ref{tab:sota-gta} and~\ref{tab:sota-synthia}. Our AutoAdapt method outperforms all the baselines. For GTA5$\rightarrow$Cityscapes, we achieve an mIoU of 50.2\%, comparable to the previous state-of-the-art method CAG \cite{zhang2019cag} (50.1\%). The other AutoAdapt methods (Auto-AdvEnt and Auto-IntraDA) outperform their baselines by 1.5\% and 1.4\% respectively. For SYNTHIA$\rightarrow$Cityscapes, we report mIoU on 13 classes (excluding ``Wall'', ``Fence'', and ``Pole'') and 16 classes. Similar observations are made. All AutoAdapt methods outperform their baselines. And our Auto-CCM achieves 53.3\% and 45.6\% mIoU on 13 classes and 16 classes respectively, both of which outperform previous state-of-the-art.

% Table generated by Excel2LaTeX from sheet 'Sheet1'
\begin{table*}[!ht]
%  \resizebox{1.0\linewidth}{!}{%
  \centering

    \begin{tabular}{l|l|l|cc|cc|c}
      \toprule
    Model & Backbone & Head  & {MMD} & ReEnt & \$Params & \#MACs& {mIoU(\%)} \\
\hline
    DeeplabV2 & Res101 [3,4,23,3]  & ASPP  & 0.78  & -0.53  & 44.6M & 380.5B & 43.8 \\
    \hline
    arch\#1 & / & -ASPP + \textbf{Attention} & 0.98  & -0.44  & 57.7M & 490.1B & 39.8 \\
    arch\#2 & / &  + \textbf{low-level} & 0.94  & -0.54  & 63.6M & 601.4B & 41.2 \\
    arch\#3 & + \textbf{SE-Layer} at Res4 & /  & 0.65  & -0.49  & 44.7M & 380.5B & 44.7 \\
    arch\#4 & / & -ASPP + \textbf{ASPP[6,18,24,36]} & 0.78   & -0.58  & 44.6M & 380.5B & 44.5 \\
    arch\#5 & -Res2 + \textbf{Res2[3]} & / & 0.77   &- 0.51  & 42.6M & 376.5B & 43.7 \\
\bottomrule

    \end{tabular}%
    % }
    \vspace{-5pt}
    \caption{Comparative studies on replacing architectures with baseline DeeplabV2. Two operations are performed to create the new architecture, replace and add resulting in 2 and 1 edit distance respectively. }
    \vspace{-10pt}
  \label{tab:arch_com}
\end{table*}%

\subsection{Ablation studies}

\label{sec:ablation}
We expect that an effective surrogate evaluation metric should be correlated with the standard metric if it could be computed (if labels were available for the target domain). Fig.~\ref{fig:ablation} shows the (negative) correlation graphically as well as through Spearman's rank correlation coefficient $\rho$ between variations of our novel evaluation metric (MMD, Entropy (Ent), and Regional Weighted Entropy (ReEnt) and the combination of them) and mIoU for GTA5$\rightarrow$Cityscapes adapted using AdvEnt. We stress that we only use the Cityscape labels to compute mIoU for this correlation analysis--we do not use the labels in our AutoAdapt framework. Fig.~\ref{fig:ablation} shows that MMD which only measures the distribution distance between source and target domains performs the worst. Combining MMD and the output structure metric Ent or ReEnt improves the correlation significantly. The proposed ReEnt by itself shows higher correlation than standard entropy. The proposed joint metric which consists of MMD and ReEnt achieves the best correlation with $\rho = -0.87$. More details of the evaluation metric can be found in the supplementary materials.

%%%Fig.~\ref{fig:ablation} presents the results of ablation studies of the proposed novel performance estimation metric. These results are for GTA5$\longrightarrow$Cityscapes with AdvEnt. We calculate the correlation between the metric (MMD, Entropy (Ent), and Regional Weighted Entropy (ReEnt) and the combination of them) and the accuracy metric mIoU. It's noted that we are not allowed to use any lables to compute the accuracy in target. Here, we assume we have labels in order to show the strong correlations between the proposed metric without label and accuracy with label. Spearman's rank correlation coefficient is used in this paper. As shown in Fig.~\ref{fig:ablation}, MMD which only measures the distribution distance between source and target domain perform worst. But when we combine MMD and the output structure metric Ent or ReEnt, we can see its improvement. Also, the proposed ReEnt shows higher correlation to accuracy compared to the regular entropy. Finally, the proposed joint metric consists of MMD and ReEnt achieved the best correlation with a p={-0.87}. More details of the performance estimation metric are shown in the supplementary materials.

 \begin{figure}[!ht]
    \centering
    \includegraphics[width=\linewidth]{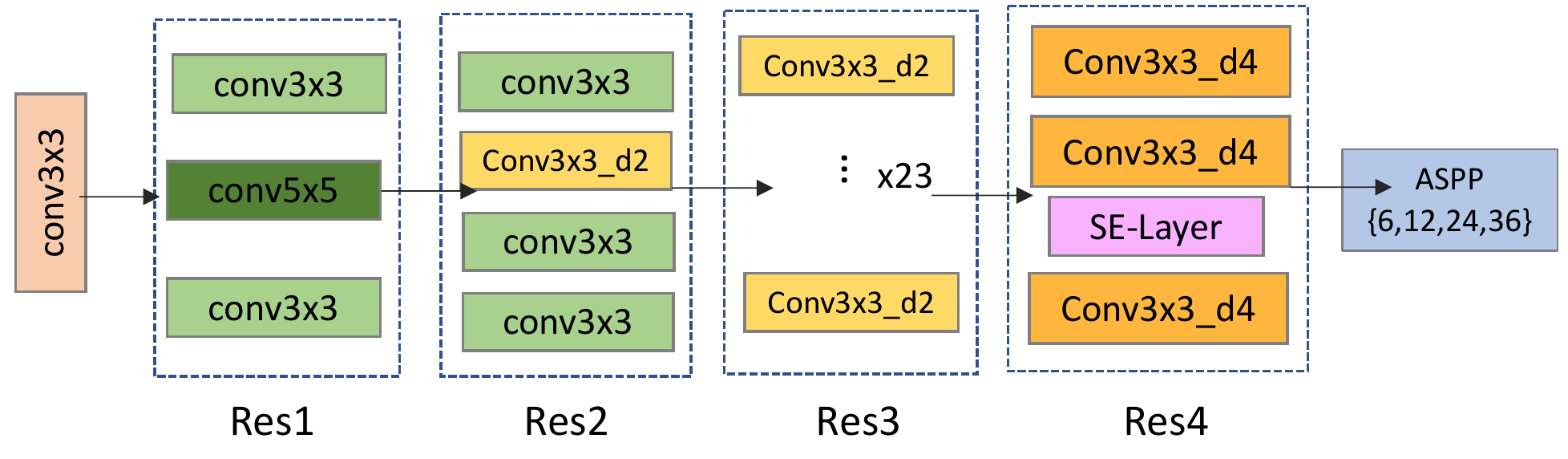}

    \caption{The architecture discovered by AutoAdapt.}

    \label{fig:res}
\end{figure}

\subsection{Architecture-level analysis on UDA}

\label{sec:arch_level_analysis}
We perform architecture-level UDA analysis in this section. Interesting observations can be made which, in some cases, contradict the observations of supervised learning. Fig.~\ref{fig:tsne} shows the TSNE embeddings of features from images from the source and target domains. These features are from the last layer of the models. Tab.~\ref{tab:arch_com} compares the results of changing only one part of the baseline Deeplabv2. Note that the ReEnt values are negative since it is a reverse entropy. Higher $|$ReEnt$|$ indicates a more even distribution corresponding to a smoother prediction. Thus the more negative ReEnt, the better.

%%%We provide a architecture-level UDA analysis in this section. Some interesting observations have been made which contradicts the regular supervised learning observations. Fig.~\ref{fig:tsne} shows the TSNE embeddings for source and target domains. We plot the TSNE embeddings calculated by features from the last layer of the model. Tab.~\ref{tab:arch_com} show the comparisons on changing only one part of baseline Deeplabv2. It's noted that all numbers are ReEnt are negative numbers, since it is a reverse entropy. Hihger $|ReEnt|$ indicates even distribution with the region showing smoother prediction and thus for ReEnt smaller the better.

\noindent \textbf{Observation 1:} As shown in Fig.~\ref{fig:tsne}, attention to spatial locations and low-level features are not beneficial for transferring knowledge or improving generalization but do result in more discriminative features as indicated by the larger number of clusters. This potentially explains why these variations are effective for supervised semantic segmentation but yield poor performance for UDA. Tab.~\ref{tab:arch_com} presents these results quantitatively (results rows 2 and 3) in that the variations result in higher MMD compared to the baseline.

%%%\noindent \textbf{Observation1:}As shown in the figure, attention on spatial locations and low-level feature are not beneficial for transferring knowledge or improving generalization but resulting in more discriminative features. This somehow explains why they are more successful in supervised semantic segmentation but yielding poor performance for UDA. Tab.~\ref{tab:arch_com} show the quantitative results (row1 and row2), they both show increased MMD compared to baseline.

\noindent \textbf{Observation 2:} Attention on channels (SE-layer) results in better feature alignment. This is likely because the operations can reweight the channel weights. Row 4 shows the model results in lower MMD which corresponds to a smaller distance between the source and target domains.

%%%\noindent \textbf{Observation2:} Attention on channels (se-layer) results in better feature alignment. That is likely because of the operations can reweight channel weights. Row 4 show the model can result in lower MMD which means a smaller distance between source and target domain.

\noindent \textbf{Observation 3:} As shown in row 5, changing the dilation rates improves the output structure (lower ReEnt value) without increasing the computational complexity. By changing the dilation rates, the model is able to generate smoother predictions with the computational burden. This provides the benefits of adding low-level feature to the architecture which is very computationally intensive but not helpful for aligning the feature distributions. Row 5 shows the model keeps the same MMD as the baseline in row 1 but improves the ReEnt.

%%%\noindent \textbf{Observation3:} Changing the dilation rates can improve the output structure with a lower ReEnt value without adding the computation complexity. By changing the dilation rates, the model is able to generate smooth prediction with having computation burdens. This provides the benefits to add low-level feature to the architecture which is very computation intensive and not helpful for aligning distribution. Row 5 shows the model keep the same MMD but improve the ReEnt.

Fig.~\ref{fig:res} shows the final model that results from our search. It has slightly more parameters but the computational complexity does not change. This model benefits mainly from adding the SE-Layer and changing the dilation rates for the ASPP based on the above observations. We conclude that the increase in parameter size is worthwhile for the improved adaptation performance. However, during search, we have the other models with less parameters and computation complexity which yield a bit worse adaptation performance compared to the final model but still outperform the baseline DeeplabV2.

%%%Fig.~\ref{fig:res} shows our final searched result which has a small amount of higher parameter size and unchanged computation complexity. This mainly benefits from adding SE-Layer and changing the dilation rates for ASPP based on the above observations. We conclude the amount of increased parameter size is worthwhile for improving the adaptation performance.

\section{Discussion and conclusion}

UDA has not experienced the kinds of improvement that supervised semantic segmentation has recently. We therefore take a different approach to UDA, one based on \textit{architecture}. We adapt NAS techniques to provide in-depth analysis on architecture for UDA. We overcome the optimization gap in NAS for UDA by developing a novel performance estimation metric. Our experimental results show that the architecture discovered using our AutoAdapt framework outperforms all the baseline UDA methods. 

We now discuss some limitations. First, the search space is relatively small compared to the supervised cases for image classification, semantic segmentation, etc. It would make sense to develop a benchmark search space for UDA in the future. Second, we only identify the challenge of the optimization gap during search. Other challenges related to the search strategy remain unsolved. For example, efficient gradient-based methods cannot be used to search architectures for UDA in semantic segmentation. Since we are the first to investigate NAS for UDA, in the future, we will try to tackle this challenge to improve the search efficiency. Third, we only partially answer the open question ``what is the role of architecture in UDA?'' Based on our observations, architectures that have various levels of transferability/generalization benefit tasks like UDA. We still feel this is a promising direction to explain the role the neural network plays in UDA and how this role might be different than in other tasks.

%%%Recently, UDA is not achieved significant improvement as done in supervised semantic segmentation in the community. We take a different perspective to tackle the problem, \textit{architecture}. We propose to adopt NAS techniques to provide in-depth analysis on architecture. We discuss the limitations in this section. First, the search space is relatively small compared to those supervised applications for image classification, semantic segmentation, etc. In the future, a benchmark search space for UDA can be proposed. Second, we only identify the challenge of optimization gap during search in this paper. Some other challenges of the search strategy remain unsolved, for example, gradient-based methods which can provide fast search can't search valuable architectures for UDA problem in semantic segmentation. Since we are the first work investigating NAS for UDA, in the future, we will try to tackle this challenge to improve the search efficiecy. Third, we partially answer the open question "what is the role of architecture for UDA?"  Based on our observation, architectures have various levels of transferability/generalization which benefits some tasks like UDA. We still feel this is a promising direction for explaining how neural network is playing its role in UDA and how it is different from the other tasks.

\label{sec:conclusion}

{\small
\bibliographystyle{ieee_fullname}
\bibliography{egbib}
}

\end{document}